\documentclass[10pt, conference, compsocconf]{IEEEtran}
\pdfoutput=1 
\IEEEoverridecommandlockouts
\usepackage{cite}
\usepackage{amsmath,amssymb,amsfonts}
\usepackage{algorithmic}
\usepackage{graphicx}
\usepackage{verbatim}
\usepackage{hyperref}
\usepackage{multirow}
\usepackage{color}
\usepackage{url}
\usepackage{textcomp}
\usepackage{xcolor}
\def\BibTeX{{\rm B\kern-.05em{\sc i\kern-.025em b}\kern-.08em
    T\kern-.1667em\lower.7ex\hbox{E}\kern-.125emX}}
\begin{document}

\title{Quantification and Analysis of Scientific Language Variation Across Research Fields\\}

\author{
	\IEEEauthorblockN{Pei Zhou, Muhao Chen, Kai-Wei Chang, Carlo Zaniolo}
	\IEEEauthorblockA{University of California, Los Angeles
		\\\{peiz, muhaochen, kwchang, zaniolo\}@cs.ucla.edu}

}

\maketitle

\begin{abstract}
Quantifying differences in terminologies from various academic domains has been a longstanding problem yet to be solved.
We propose a computational approach for analyzing linguistic variation among scientific research fields by capturing the semantic change of terms based on a neural language model.
The model is trained on a large collection of literature in five computer science research fields, for which we obtain field-specific vector representations for key terms, and global vector representations for other words. 
Several quantitative approaches are introduced to identify the terms whose semantics have drastically changed, or remain unchanged across different research fields.
We also propose a metric to quantify the overall linguistic variation of research fields.
After quantitative evaluation on human annotated data and qualitative comparison with other methods, we show that our model can improve cross-disciplinary data collaboration by identifying terms that potentially induce confusion during interdisciplinary studies.

\end{abstract}
\section{Introduction}

The usage of language always varies among people with different backgrounds.
When it comes to scientific literature, linguistic variation commonly exists across different scientific fields, among which we often see that scholars with varied backgrounds of knowledge use the same terms to express entirely different meanings.
For this paper, we mainly consider different research fields within one subject (Computer Science) such as {\em Natural Language Processing} (NLP) and {\em Computer Networks and Communication} (Comm) instead of different subjects.
This is because research fields within one subject usually have more shared terms whose semantic changes may lead to ambiguity \cite{r2}.
Consider the term {\em alignment}, which often refers to the matching of signals in the field of computer communications, is however more typically related to translation of words or sentences in NLP research.
Other representative examples include terms such as {\em embedding}, {\em semantic}, and {\em grid}, etc.
Against this issue, quantified analysis of linguistic variation for scientific terms benefits with clearer understanding of concept expressions in different scientific fields, and reducing confusions for interdisciplinary communications.
Moreover, such computational methods can also demonstrate the divergence of overall language trends among research fields.\par
\begin{figure}
	\includegraphics[width=\columnwidth]{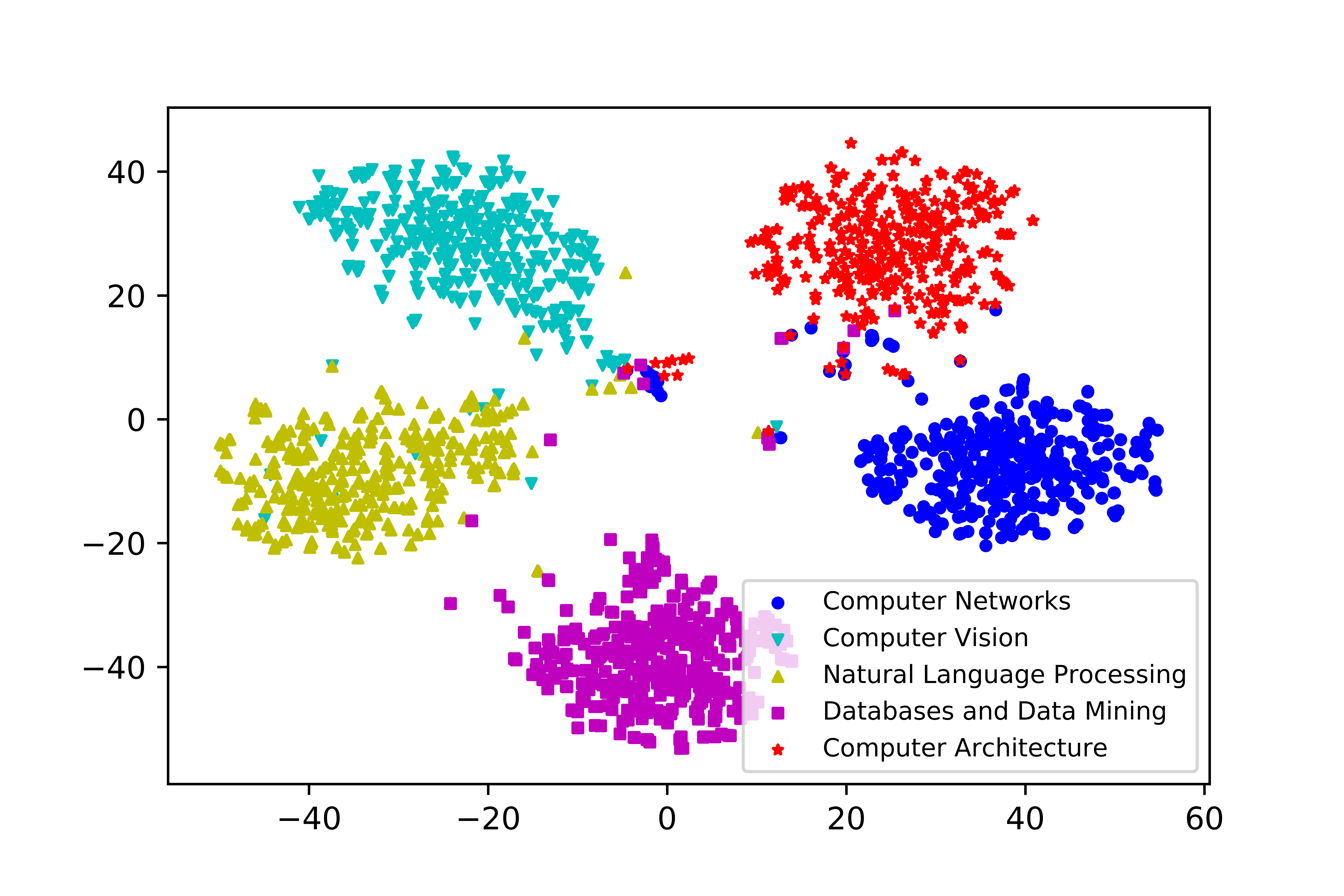}
	\vspace{-1em}
	\centering
	\caption{Visualization of scientific term embeddings using our model after projecting on two dimensional space.}~\label{fig:w2v_vis}
\end{figure}
While computational analysis methods of linguistic variation have attracted much attention recently, past work mainly focused on geographic, temporal, and social aspects of languages~\cite{hu2017world,kulkarni2016freshman,kulkarni2015statistically,kim2014temporal,bamman2014distributed,bamman2014gender,o2010discovering,zeng2017socialized,eisenstein2014diffusion,eisenstein2010latent}.
Studies on these aspects mainly focus on differentiating word representations that change dramatically in the corresponding aspect for the entire vocabulary.
However, existing work has not really tackled field variation of the language that modifies the meanings of scientific terms across these research fields, nor have they utilized these terms to reveal the general inter-field divergence of the language.

In this paper, we propose a computational method to analyze the linguistic variation in scientific research fields based on the quantification of semantic changes of crucial scientific terms.
We extend the neural language model~\cite{mikolov2013efficient,mikolov2013distributed} with a partially-localized mechanism to capture field-specific embeddings for frequent key terms, and preserve the global representations for other general words.
Fig.~\ref{fig:w2v_vis} shows the the embedded word vectors for key terms in five domains after reducing the dimension of the embeddings, and we can see clear separation between domains.
We also propose metrics to quantify the linguistic variation for scientific fields based on the semantic distances of key terms.
The proposed approach hence helps keep track of the variation of term expressions, and provides evidences for speculations about the diversity of language usage across different research areas.\par


\section{Modeling}
We begin with the formalization of the corpus $D$, which is a sequence of words.
We use $F=\left \{ f_1, f_2, ..., f_m \right \}$ to denote the set of fields.
$D$ is partitioned into $m$ disjoint field-specific corpora, i.e., $D=\bigcup_{f_i\in F} D_{f_i}$.
We use $V=V_{G}\cup V_{F}$ to denote the vocabulary of words, for which $V_G=\left \{ w_{t} \right \}$ is the global vocabulary, and $V_F=\bigcup_{f_i\in F}V_{f_i}$, which is disjoint with $V_G$, is the vocabulary of field-specific terms, where $V_{f_i}=\left \{ w_{t}^{f_i \in F} \right \}$.
In practice, $V_F$ can be predefined by selecting the most frequent terms from titles.
Note that for every $f_{i}, f_{j} \in F$, $w_{t}^{f_{i}}$ and $w_{t}^{f_{j}}$ represent one term $w_t$ that occurs in two field-specific corpora, and thus have two different embeddings.
We establish two disjoint sets of vocabularies in order to localize embeddings for key terms and keep generic embeddings for other sentence elements.
For a word $w_t \in D_{f_i}$, $C_{f_i} (w_{t})=\left \{ w_{t-s}, ..., w_{t-1}, w_{t+1}, ..., w_{t+s} \right \}$ is the context of $w_t$, where $s$ is half of the context size, and
each word in $C_{f_i} (w_{t})$ belongs to either $V_G$ or $V_{f_i}$.

\subsection{Neural Language Model}
We extend the 
CBOW model~\cite{mikolov2013efficient} to capture both global embeddings of words and field-specific embeddings of scientific terms.
The training objective of the model is to learn word vector representations that maximize the log likelihood of the word and term given its context that is specified in different fields:
\begin{equation*}
J=\sum_{i}^{m} \sum_{w_{t}\in V}\log P(w_t\mid C_{f_i}(w_t)),
\end{equation*}
for which the conditional likelihood of word $w_t$ over a field-specific context is defined as:
\begin{equation*}
P(w_t\mid C_{f_i}(w_t)) = \frac{\exp(\mathbf{w}_t^{\top}\cdot vec(C_{f_i}(w_t)))}{\sum_{w\in D_{f_i}}\exp(\mathbf{w}_t^{\top}\cdot vec(C_{f_i}(w)))},
\end{equation*}
where $\mathbf{w}_t$ is the embedding of word $w_t$, and context $C_{f_i}(w_t)$ is represented as the mean of contained word embeddings, i.e., $vec(C_{f_i}(w_t)) = \frac{1}{2s} \sum_{w_{i}\in C_{f_i(w_t)}}\mathbf{w}_i$.
Following~\cite{mikolov2013efficient}, we adopt batched negative sampling to obtain a computationally efficient approximation of log likelihood. 
\subsection{Quantification Approaches}
In this section, we present several methods to quantify the linguistic variation for terms and research fields.
\subsubsection{Term Variation}
We consider both Cosine Similarity: $\mathrm{cos}\left ( w_i,w_j \right )=\frac{\mathbf{w_i}\cdot \mathbf{w_j}}{\left \| \mathbf{w_i} \right \|_{2}\left \| \mathbf{w_j} \right \|_{2}}$ and Jaccard Similarity for quantitative analysis.
To get the Jaccard Similarity, we find the most similar $k$ words or terms using cosine similarity for a term $w_{t}^{f}$ in $V_F$, based on which the variation between the term in two fields (i.e. $w_{t}^{f_i}, w_{t}^{f_j}$) is calculated with Jaccard Distance. This similarity metric aims to capture the second order similarity for different terms. We consider the set of $k$ most similar words in order to capture the latent semantic of the term:
\begin{equation*}
Dis^k(w_{t}^{f_i},w_{t}^{f_j})=\frac{\left |W^{k}_i\cup W^{k}_j \right |-\left | W^{k}_i\cap W^{k}_j \right |}{\left | W^{k}_i\cup W^{k}_j \right |}.
\end{equation*}
$W^{k}_i$ thereof, indicates the set of $k$ most similar words considering its cosine similarities with $w_i$, i.e., $W^{k}_i=\left \{ w_t \mid \mathrm{cos}(w_t,w_i) \in \mathrm{topk}(\bigcup_{w \in V}\left \{ \mathrm{cos}(w_i,w) \right \})\right \}$, 
for which $\mathrm{topk}(\bigcup_{w \in V}\left \{ \mathrm{cos}(w_i,w) \right \})$ denotes $k$ most similar words to $w_i$, and $J(W^{k}_i, W^{k}_j)=\frac{\left | W^{k}_i \cap W^{k}_j\right |}{\left |W^{k}_i \cup W^{k}_j  \right |}$ is the Jaccard Index to measure the similarity of two sets.
Intersection thereof is calculated by aggregating the cosine similarities of the words that appear in $W^{k}_i$ and $W^{k}_j$, and union is calculated by summing up the cosine similarities thereof.\par
In practice, a higher $k$ should cause $Dis^k$ to more comprehensively measure the variation of a term by considering the semantic distances of more related words in each field, which however requires more computational cost.

\subsubsection{Field Variation}


We quantify the overall linguistic variation of research fields $f_i, f_j$ based on the field-specific term vocabularies $V_{f_i}, V_{f_j}$.
For each term $w^{f_i}_t \in V_{f_i}$, we first find the set of $k$ most similar words $W_{t,f_i}^{k}$.
Then the subset of $W_{t,f_i}^{k}$ that contains terms of the second field $f_j$ is obtained as $S_{t,f_i,f_j}^k= \left \{ w_{t}^{f_j} \mid  w_{t}^{f_j} \in (W_{t,f_i}^{k}\bigcap V_{f_j})\right \}$.
Based on that, we aggregate a term-to-field semantic similarity measure $sim^k(w_{t}^{f_i}, f_j)$, i.e., 
\begin{equation*}
\resizebox{\columnwidth}{!}{$sim^k(w_{t}^{f_i}, f_j)=\frac{\sum_{w\in S_{t,f_i,f_j}^k} f(w^{f_i}_{t}) \cdot \mathrm{cos}(w^{f_i}_{t}, w) \cdot f(w)}{\left | S_{t,f_i,f_j}^k \right |},$}
\end{equation*}
where $f(w^{f_i}_t)$ is the frequency of word $w^{f_i}_t$, i.e. number of occurrence of word $w^{f_i}_t$ in $D_{f_i}$ divided by the total number of words in $D_{f_i}$.\par

Then the semantic distance between two fields $f_i, f_j$ is defined based on the aggregated term-to-field similarities for all words in $V_{f_i}$, for which we apply normalized exponential scaling~\cite{mondain1998polymer} to signify the differences of measures for different fields. $\lambda=e-1$ thereof is a normalizing constant. Thus we have:\par
\begin{equation*}
\resizebox{\columnwidth}{!}{$FieldDis^k(f_i,f_j)=\frac{\exp(1-\sum_{w^{f_i}_t \in V_{f_i}} sim^k(w^{f_i}_t,f_j))-1}{\lambda}.$}
\end{equation*}

\section{Evaluation}
In this section, 
we evaluate the proposed approach for analyzing term and field-level linguistic variation based on a large collection of scientific literature.
We collect human annotations on term variation and compare the results with separately-trained CBOW, GEODIST \cite{kulkarni2016freshman}, and our model.\par

\subsection{Dataset and Model Configuration}
We train our model on research papers in five fields of computer science, including NLP, {\em Computer Vision} (CV), {\em Databases and Data Mining} (DBDM), {\em Computer Architecture} (Arch), and {\em Computer Networks and Communication} (Comm).
We select around ten mainstream conferences and journals for each field based on the Google Scholar Publication Ranking List \footnote{\url{https://scholar.google.com/citations?view_op=top_venues&hl=en&vq=eng}}, and correspondingly collected the papers in these venues that are published in the past seven years.
The corpora contain the plaintext contents of over 56K academic papers.\par

We populate $V_F$ using the article titles, where the stop words are removed and 200 most frequent terms are selected for each field.
Since the title should reflect the main topic of each paper, this word collection naturally contains a set of very popular scientific terms. 
We set the dimensionality of embeddings to be 100, the size of the contexts $s$ to be 24, and the negative sample size to be 5.

\subsection{Variation of Terms}
\begin{table}[t]
\begin{minipage}[t]{0.55\linewidth}
\caption{\label{most_varied} Terms with most significant semantic change.}
\setlength\tabcolsep{1pt}
\centering
{
\begin{tabular}{|c|c|c|}
\hline
\textbf{Term} & \textbf{Avg. $Dis^{k}$} & \textbf{Highest $Dis^{k}$ and fields} \\ \hline
\textit{alignment}        & 0.9215                   & 0.9634 (NLP-Comm)       \\ \hline
\textit{relations}        & 0.8923                   & 0.9654 (NLP-Arch)      \\ \hline
\textit{translation}      & 0.8905                   & 0.9773 (NLP-Arch)  \\ \hline
\textit{mapping}          & 0.8819                   & 0.9817 (NLP-Arch)      \\ \hline
\textit{embedding}        & 0.8780                   & 0.9533 (NLP-Comm)         \\ \hline
\textit{embed}            & 0.8675                   & 0.9785 (NLP-Arch)     \\ \hline
\textit{pattern}          & 0.8670                   & 0.9804 (CV-Arch)        \\ \hline
\textit{feature}          & 0.8621                   & 0.9667 (NLP-Arch)     \\ \hline
\textit{grid}             & 0.8509                   & 0.9445 (NLP-CV)      \\ \hline
\textit{semantic}         & 0.8396                   & 0.9679 (NLP-Arch)      \\ \hline
\end{tabular}
}
\end{minipage}
\hfill
\begin{minipage}[t]{0.33\linewidth}
\caption{\label{least_varied} Terms with least semantic change.}
\centering
\setlength\tabcolsep{1pt}
{
\begin{tabular}{|c|c|}
\hline
\textbf{Term} & \textbf{Avg. $Dis^{k}$} \\ \hline
\textit{linux}             & 0.4721                   \\ \hline
\textit{cryptography}      & 0.5283                   \\ \hline
\textit{multicore}         & 0.5309                   \\ \hline
\textit{imaging}           & 0.5352                   \\ \hline
\textit{intel}             & 0.5452                   \\ \hline
\textit{multiprocess}   & 0.5512                   \\ \hline
\textit{wireless}          & 0.5566                   \\ \hline
\textit{hardware}          & 0.5618                   \\ \hline
\textit{telecom}           & 0.5639                   \\ \hline
\textit{server}            & 0.5664                   \\ \hline
\end{tabular}
}
\end{minipage}
\end{table}
\begin{table}[t]
\centering
\caption{\label{sim_words} Most similar words of some terms in Table~\ref{most_varied} and \ref{least_varied}.}
\resizebox{\columnwidth}{!}{
	\setlength\tabcolsep{1pt}
	\tiny
	\begin{tabular}{|c|c|c|c|c|c|}
	\hline
	\multirow{2}{*}{\textbf{Term}} & \multicolumn{5}{c|}{\textbf{Most Similar Words Per Field}}                                                                                                                                                                                                                                                                                                                                               \\ \cline{2-6}
	                               & \textbf{NLP}                                                                 & \textbf{CV}                                                                        & \textbf{DBDM}                                                           & \textbf{Arch}                                                           & \textbf{Comm}                                                             \\ \hline
	\textit{alignment}             & \begin{tabular}[c]{@{}c@{}}translation\\ bilingual\\ parallel\end{tabular}   & \begin{tabular}[c]{@{}c@{}}facial\\ shape\\ localization\end{tabular}              & \begin{tabular}[c]{@{}c@{}}matching\\ alignments\\ mapping\end{tabular} & \begin{tabular}[c]{@{}c@{}}sequences\\ binary\\ parallel\end{tabular}   & \begin{tabular}[c]{@{}c@{}}movr\\nlos\\ phasedarray\end{tabular}         \\ \hline
	\textit{embedding}             & \begin{tabular}[c]{@{}c@{}}embeddings\\ representation\\ vector\end{tabular} & \begin{tabular}[c]{@{}c@{}}embed\\ similarity\\ sparsecoded\end{tabular}           & \begin{tabular}[c]{@{}c@{}}embeddings\\ structure\\ space\end{tabular}  & \begin{tabular}[c]{@{}c@{}}embed\\ tree\\ bitsequence\end{tabular}      & \begin{tabular}[c]{@{}c@{}}embedded\\ visual\\ domain\end{tabular}        \\ \hline
	\textit{semantic}              & \begin{tabular}[c]{@{}c@{}}semantics\\ syntactic\\ lexical\end{tabular}      & \begin{tabular}[c]{@{}c@{}}attributes\\ segmentation\\ representation\end{tabular} & \begin{tabular}[c]{@{}c@{}}ontology\\ semantics\\ web\end{tabular}      & \begin{tabular}[c]{@{}c@{}}semantics\\ tracking\\ approach\end{tabular} & \begin{tabular}[c]{@{}c@{}}structured\\ relations\\ indexing\end{tabular} \\ \hline
	\textit{linux}                 & \begin{tabular}[c]{@{}c@{}}workstation\\ intel\\ unix\end{tabular}           & \begin{tabular}[c]{@{}c@{}}intel\\ geforce\\ workstation\end{tabular}              & \begin{tabular}[c]{@{}c@{}}intel\\ quadcore\\ operating\end{tabular}    & \begin{tabular}[c]{@{}c@{}}kernel\\ software\\ filesystem\end{tabular}  & \begin{tabular}[c]{@{}c@{}}kernel\\ userspace\\ software\end{tabular}     \\ \hline
	\textit{cryptography}         & \begin{tabular}[c]{@{}c@{}}public-key\\ secrets\\ backroom\end{tabular}      & \begin{tabular}[c]{@{}c@{}}secure\\ public-key\\ encryption\end{tabular}           & \begin{tabular}[c]{@{}c@{}}crypto\\ secure\\ encryption\end{tabular}    & \begin{tabular}[c]{@{}c@{}}public-key\\ crypto\\ secure\end{tabular}    & \begin{tabular}[c]{@{}c@{}}crypto\\ public-key\\ secure\end{tabular}      \\ \hline
	
	\end{tabular}
}
\end{table}

To evaluate the semantic change of a term $w_t$, we calculate the $Dis^k$ with the field-specific embeddings of this term, i.e. $\mathbf{w}^{f_i}_t$ and $\mathbf{w}^{f_j}_t$.
Specifically, we analyze the $Dis^k$ of one term for in total ten pairs of fields, and for which average of their distances to quantify the overall variation of the term.
We set $k$ to be 10,000 during the evaluation, which seeks to aggregate semantics from a large neighborhood of each term.
Results reported in Table~\ref{most_varied} show the ten terms with the overall most significant semantic changes, and between which fields such changes happen the most, while those in Table~\ref{least_varied} show the ten terms with the least semantic change.
The model detects that the meanings of terms like \textit{alignment} and \textit{embedding} should vary a lot across different fields, whereas those of terms like \textit{hardware} and \textit{server} are expected to be consistent.
Table~\ref{sim_words} presents the most similar words of some terms from Table~\ref{most_varied} and Table~\ref{least_varied} in each of the five field.
It is noteworthy that, such terms with a high $Dis^k$ indeed have very different meanings and usages across different fields.

\subsection{Comparison on Human Annotated Data}
\begin{table}[t]
\setlength\tabcolsep{1pt}
\caption{\label{comparison} Comparison with Human Annotations.}
\centering
\begin{tabular}{|c|c|c|c|}
\hline
Methods           & $\rho_{Pearson}$ & nDCG at rank 30 &nDCG at rank 50 \\ \hline
Separate CBOW--JS & 0.492               & 0.753           & 0.839           \\ \hline
GEODIST--Cosine   & -0.488              & 0.115           & 0.406           \\ \hline
Our Model--Cosine     & 0.777               & 0.815           & \textbf{0.858}  \\ \hline
Our Model--JS & \textbf{0.779}      & \textbf{0.817}  & \textbf{0.858}  \\ \hline
\end{tabular}
\end{table}

To quantitatively evaluate the term variation found by our model, we randomly choose 50 terms and instruct a group of 30 Computer Science PhD students to annotate them as semantically varied (+1) or not (-1), across the five research fields.
We make sure each single term is annotated by at least five PhD students, sum their annotations up, and divide the summed number by 5 to get the annotated term variation.
We collect all annotation for 50 terms and compare with the corresponding variation returned from different methods as shown in Table~\ref{comparison}.
As a baseline, separately-trained CBOW model on each of the five domains is considered, and the variation is calculated using Jaccard Similarity mentioned before.
We also ran the GEODIST model introduced by \cite{kulkarni2016freshman} on the corpus for comparison.
Then we test the performances on our model using both Cosine Similarity and Jaccard Similarity as similarity measures.
For metrics, we use Pearson Correlation ($\rho_{Pearson}$) \cite{galton1886regression} and Normalized Discounted Cumulative Gain (nDCG) at rank 10 and 50 \cite{jarvelin2002cumulated}.
From the results, we can see that our model performs better than baselines for all three metrics.
For Pearson Correlation ($\rho_{Pearson}$), the improvement is more significant and the calculated variation has a linear correlation with the ground truth annotation.
GEODIST fails to differentiate the semantic variation of terms across different groups of literature with selected terms, thus obtains a negative correlation and a low nDCG score.
We hypothesis that it is because for the research field corpus, the language style of individual paper in the same field varies greatly and it is a more noisy corpus than the Google Book Ngrams corpus they used.
Our model works well with the noisy data and also outperforms the separately-traind CBOW baseline.
We also observe that using Jaccard similarity improves the results, showing that second order similarity metrics are arguably better than first order metrics.

\subsection{Variation of Fields}


\begin{figure}
  \includegraphics[width=0.76\columnwidth]{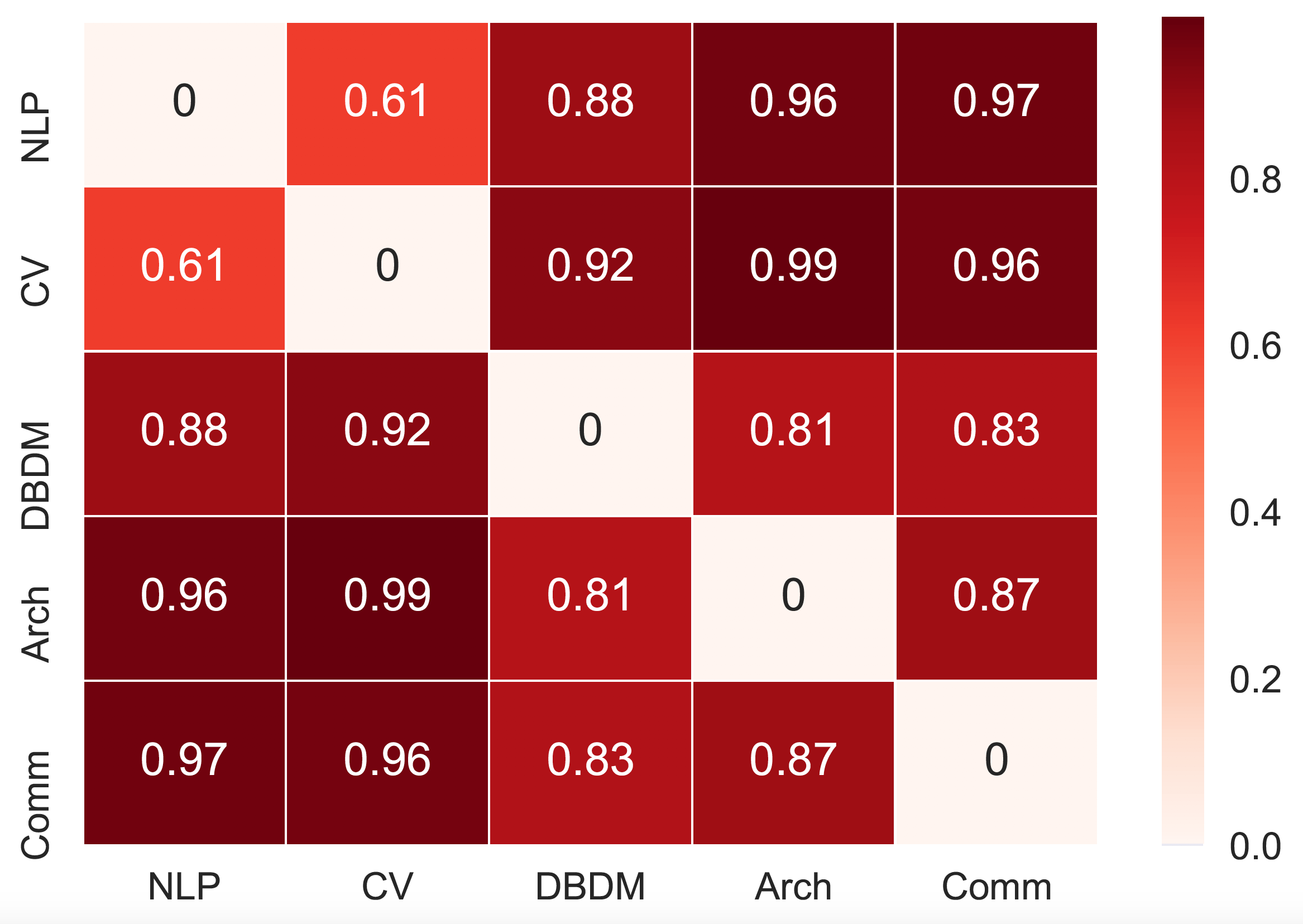}
  \vspace{-1em}
  \centering
  \caption{Heatmap of field distances.}~\label{fig:dis_fields}
  \vspace{-2em}
\end{figure}

We now analyze the overall linguistic variation of different research fields based on $FieldDis^k$, for which we set $k$ as 10,000.
The results are shown in the heatmap of Fig.~\ref{fig:dis_fields}, in which darker colors indicate larger overall semantic changes (i.e. higher $FieldDis^k$).
The results indicate that the language usages between NLP and CV are considered as most similar.
This is explainable because NLP and CV are both AI-related fields and share lots of techniques in research studies.
For DBDM, corresponding linguistic variation from other fields is more significant.
This can be explained by the fact that, on one hand, data mining tasks employ many shared statistical techniques in AI research as well, and a portion of modern database research in distributed scenarios also shares much common knowledge with computer and network systems.
On the other hand, there is still much distinction from DBDM to the rest involved fields.
Another observation is that the language usages are considered as most dissimilar from the fields of NLP and CV to the fields of Arch and Comm.
This conclusion is congruent to the evaluation results for terms in Table~\ref{most_varied}, where distances for these fields are the highest for terms with high semantic changes.
This is because AI research topics are often considered distant from computer system-related research fields like Arch and Comm, which explains why the $FieldDis^k$ measures are significant from any of the former to that of the later.
Besides, we discover that the semantic change between Arch and Comm is also relatively significant.\par


\section{Related Work}
In this section, we discuss related work on computational methods for linguistic variation.
Recent work has paid much attention to using neural language models for corresponding analysis, which are respectively based on two lines of word representation mechanisms.\par

\textbf{Diffusion-based.} The diffusion-based mechanism categorizes the corpora by different context scenarios, and obtains differentiated representations for words in each context scenario.
Corresponding methods have been used to analyze the semantics of words that vary according to different time periods or social groups \cite{kim2014temporal,kulkarni2015statistically,eisenstein2014diffusion,hu2017world}.
Our work is close to these methods, but instead of differentiating all words, we employs a partially-localized mechanism by localizing embeddings for a portion of important terms, while keep the universal representations for other generic words.
This is due to that our task naturally focuses on measuring the change of frequent scientific terms that contribute mostly to the linguistic variation among scientific research fields.\par
\textbf{Bias-based.} Other work adopts the bias-based representation mechanism, which uniformly represents all words, and overlays a scenario-specific bias vector to words that appear in the corresponding context scenario.
Exemplarily, \cite{kulkarni2016freshman} captures the biases between UK and US English based on corresponding literatures, and \cite{bamman2014distributed} induces such biases on social media corpora that are tagged with more fine-grained geo-locations.
Similarly, a user-specific bias is utilized by \cite{zeng2017socialized}.
While each context scenario often applies the same bias to its words, we prefer the diffusion-based representations due to that our task requires the semantic changes to be captured distinctively for the terms in the same field.\par

Besides, we also provide quantitaive evaluation for the task of analyzing scientific language variation for different domains.
Previous work has mainly showed qualitative analysis on language variation geographically or temporally.
We argue that by comparing with human annotations, our model can capture the semantic change for cross-domain data.
Moreover, we leverage the word-level quantification to measure the overall semantic differences of research fields.

\section{Conclusion}
In this paper we introduced a computational model for analyzing scientific linguistic variation by research fields.
The model was trained on a large collection of literature from five computer science research fields to obtain partially-localized representations for terms.
A series of metrics were provided to quantify the semantic change on both term and field level.
We evaluate the term variation found by our model by comparing with human annotated data and sevearl baselines and show that our model captures the term variation most accurately.
We believe that with automatically detected term variation, confusion during interdisciplinary communications is reduced, and the goal of better cross-domain data collaboration is achieved.


\bibliographystyle{IEEEtran}
\bibliography{reference}

\begin{thebibliography}{10}
\providecommand{\url}[1]{#1}
\csname url@samestyle\endcsname
\providecommand{\newblock}{\relax}
\providecommand{\bibinfo}[2]{#2}
\providecommand{\BIBentrySTDinterwordspacing}{\spaceskip=0pt\relax}
\providecommand{\BIBentryALTinterwordstretchfactor}{4}
\providecommand{\BIBentryALTinterwordspacing}{\spaceskip=\fontdimen2\font plus
\BIBentryALTinterwordstretchfactor\fontdimen3\font minus
  \fontdimen4\font\relax}
\providecommand{\BIBforeignlanguage}[2]{{%
\expandafter\ifx\csname l@#1\endcsname\relax
\typeout{** WARNING: IEEEtran.bst: No hyphenation pattern has been}%
\typeout{** loaded for the language `#1'. Using the pattern for}%
\typeout{** the default language instead.}%
\else
\language=\csname l@#1\endcsname
\fi
#2}}
\providecommand{\BIBdecl}{\relax}
\BIBdecl

\bibitem{r2}
R.~Kittredge and J.~Lehrberger, \emph{Sublanguage: Studies of language in
  restricted semantic domains}.\hskip 1em plus 0.5em minus 0.4em\relax Walter
  de Gruyter, 1982.

\bibitem{hu2017world}
T.~Hu, R.~Song, P.~Ding \emph{et~al.}, ``A world of difference: Divergent word
  interpretations among people,'' in \emph{ICWSM}, 2017.

\bibitem{kulkarni2016freshman}
V.~Kulkarni, B.~Perozzi, and S.~Skiena, ``Freshman or fresher? quantifying the
  geographic variation of language in online social media.'' in \emph{ICWSM},
  2016, pp. 615--618.

\bibitem{kulkarni2015statistically}
V.~Kulkarni, R.~Al-Rfou, B.~Perozzi, and S.~Skiena, ``Statistically significant
  detection of linguistic change,'' in \emph{Proceedings of the 24th
  International Conference on World Wide Web}.\hskip 1em plus 0.5em minus
  0.4em\relax International World Wide Web Conferences Steering Committee,
  2015, pp. 625--635.

\bibitem{kim2014temporal}
Y.~Kim, Y.-I. Chiu, K.~Hanaki, D.~Hegde, and S.~Petrov, ``Temporal analysis of
  language through neural language models,'' \emph{ACL 2014}, p.~61, 2014.

\bibitem{bamman2014distributed}
D.~Bamman, C.~Dyer, N.~A. Smith \emph{et~al.}, ``Distributed representations of
  geographically situated language,'' in \emph{Proceedings of the 52nd Annual
  Meeting of the Association for Computational Linguistics}, 2014.

\bibitem{bamman2014gender}
D.~Bamman, J.~Eisenstein, and T.~Schnoebelen, ``Gender identity and lexical
  variation in social media,'' \emph{Journal of Sociolinguistics}, vol.~18,
  no.~2, pp. 135--160, 2014.

\bibitem{o2010discovering}
B.~O'Connor, J.~Eisenstein, E.~P. Xing \emph{et~al.}, ``Discovering demographic
  language variation,'' in \emph{MLSC}, 2010.

\bibitem{zeng2017socialized}
Z.~Zeng, Y.~Yin, Y.~Song, and M.~Zhang, ``Socialized word embeddings,'' in
  \emph{Proceedings of the 26th International Joint Conference on Artificial
  Intelligence}.\hskip 1em plus 0.5em minus 0.4em\relax AAAI Press, 2017, pp.
  3915--3921.

\bibitem{eisenstein2014diffusion}
J.~Eisenstein, B.~O'Connor, N.~A. Smith, and E.~P. Xing, ``Diffusion of lexical
  change in social media,'' \emph{PloS one}, vol.~9, no.~11, p. e113114, 2014.

\bibitem{eisenstein2010latent}
------, ``A latent variable model for geographic lexical variation,'' in
  \emph{Proceedings of the 2010 Conference on Empirical Methods in Natural
  Language Processing}.\hskip 1em plus 0.5em minus 0.4em\relax Association for
  Computational Linguistics, 2010, pp. 1277--1287.

\bibitem{mikolov2013efficient}
T.~Mikolov, K.~Chen, G.~Corrado \emph{et~al.}, ``Efficient estimation of word
  representations in vector space,'' \emph{ICLR Workshops Track}, 2013.

\bibitem{mikolov2013distributed}
T.~Mikolov, I.~Sutskever, K.~Chen, G.~S. Corrado, and J.~Dean, ``Distributed
  representations of words and phrases and their compositionality,'' in
  \emph{Advances in neural information processing systems}, 2013, pp.
  3111--3119.

\bibitem{mondain1998polymer}
O.~Mondain-Monval, A.~Espert, P.~Omarjee, J.~Bibette, F.~Leal-Calderon,
  J.~Philip, and J.-F. Joanny, ``Polymer-induced repulsive forces: Exponential
  scaling,'' \emph{Physical review letters}, vol.~80, no.~8, p. 1778, 1998.

\bibitem{galton1886regression}
F.~Galton, ``Regression towards mediocrity in hereditary stature.'' \emph{The
  Journal of the Anthropological Institute of Great Britain and Ireland},
  vol.~15, pp. 246--263, 1886.

\bibitem{jarvelin2002cumulated}
K.~J{\"a}rvelin and J.~Kek{\"a}l{\"a}inen, ``Cumulated gain-based evaluation of
  ir techniques,'' \emph{ACM Transactions on Information Systems (TOIS)},
  vol.~20, no.~4, pp. 422--446, 2002.

\bibitem{blei2003latent}
D.~M. Blei, A.~Y. Ng, and M.~I. Jordan, ``Latent dirichlet allocation,''
  \emph{Journal of machine Learning research}, vol.~3, no. Jan, pp. 993--1022,
  2003.

\end{thebibliography}


\begin{thebibliography}{00}
\bibitem{b1} G. Eason, B. Noble, and I. N. Sneddon, ``On certain integrals of Lipschitz-Hankel type involving products of Bessel functions,'' Phil. Trans. Roy. Soc. London, vol. A247, pp. 529--551, April 1955.
\bibitem{b2} J. Clerk Maxwell, A Treatise on Electricity and Magnetism, 3rd ed., vol. 2. Oxford: Clarendon, 1892, pp.68--73.
\bibitem{b3} I. S. Jacobs and C. P. Bean, ``Fine particles, thin films and exchange anisotropy,'' in Magnetism, vol. III, G. T. Rado and H. Suhl, Eds. New York: Academic, 1963, pp. 271--350.
\bibitem{b4} K. Elissa, ``Title of paper if known,'' unpublished.
\bibitem{b5} R. Nicole, ``Title of paper with only first word capitalized,'' J. Name Stand. Abbrev., in press.
\bibitem{b6} Y. Yorozu, M. Hirano, K. Oka, and Y. Tagawa, ``Electron spectroscopy studies on magneto-optical media and plastic substrate interface,'' IEEE Transl. J. Magn. Japan, vol. 2, pp. 740--741, August 1987 [Digests 9th Annual Conf. Magnetics Japan, p. 301, 1982].
\bibitem{b7} M. Young, The Technical Writer's Handbook. Mill Valley, CA: University Science, 1989.
\end{thebibliography}

\end{document}